\documentclass[11pt]{article}

\usepackage[final]{acl}
\usepackage{times}
\usepackage{latexsym}
\usepackage[T1]{fontenc}
\usepackage[utf8]{inputenc}
\usepackage{microtype}
\usepackage{inconsolata}
\usepackage{graphicx}
\usepackage{booktabs}
\usepackage{amsmath}
\usepackage{amssymb}
\usepackage{multirow}
\usepackage{enumitem}
\usepackage{xcolor}
\usepackage{listings}
\usepackage{float}
\usepackage{url}
\usepackage{algorithm}
\usepackage{algpseudocode}
\usepackage{placeins}
\setlist{noitemsep,topsep=1pt,parsep=0pt,partopsep=0pt,leftmargin=*}
\newcommand\blfootnote[1]{%
  \begingroup
  \renewcommand\thefootnote{}\footnote{#1}%
  \addtocounter{footnote}{-1}%
  \endgroup
}

\title{LivingArena: Do LLMs Know What Other LLMs Don't?\\Peer-Probing as Scalable Evaluation}

\author{%
  Xingyu Chen \\
  Shanghai Jiao Tong University, Tencent \\
  \texttt{galaxychen@sjtu.edu.cn}
  \And
  Rui Wang$^{*}$ \\
  Shanghai Jiao Tong University \\
  \texttt{wangrui12@sjtu.edu.cn}
  \AND
  Zhaopeng Tu$^{*}$ \\
  Tencent \\
  \texttt{tuzhaopeng@gmail.com}
  \And
  Liefeng Bo \\
  Tencent \\
  \texttt{liefengbo@gmail.com}}

\begin{document}
\maketitle
\blfootnote{$^{*}$~Corresponding authors.}

\begin{abstract}
Fixed benchmarks are costly to renew and cannot adapt their questions to model-specific failures. We ask whether LLMs can instead discover one another's weaknesses and turn those observations into an evaluation process. To study this question, we introduce \textbf{LivingArena}, an automated peer-probing framework in which models take turns testing one another. Using the interaction history, each questioner identifies potential weaknesses of its opponent and constructs targeted, verifiable questions to probe them. A 3,600-round tournament of ten models reveals a clear role asymmetry: strong answerers are not always reliable questioners, because they may generate internally inconsistent tests or fail to verify their own reference answers. After a questioner exposes an answerer's failure, it is more likely to pursue the same capability domain, while the answerer's weakness recurs on independently generated questions, including questions written by different models. These findings show that peer probing can reveal persistent model-specific weaknesses while separately evaluating answering and reliable test construction. By automating this process and allowing test difficulty to evolve with model capabilities, LivingArena provides a ``living'' benchmark for model development, red-teaming, and capability-aware multi-agent coordination. We publicly release our code.\footnote{\url{https://github.com/galaxyChen/LivingArena}}
\end{abstract}

\begin{figure}[!t]
  \vspace{1.4em}
  \centering
  \includegraphics[width=0.90\columnwidth]{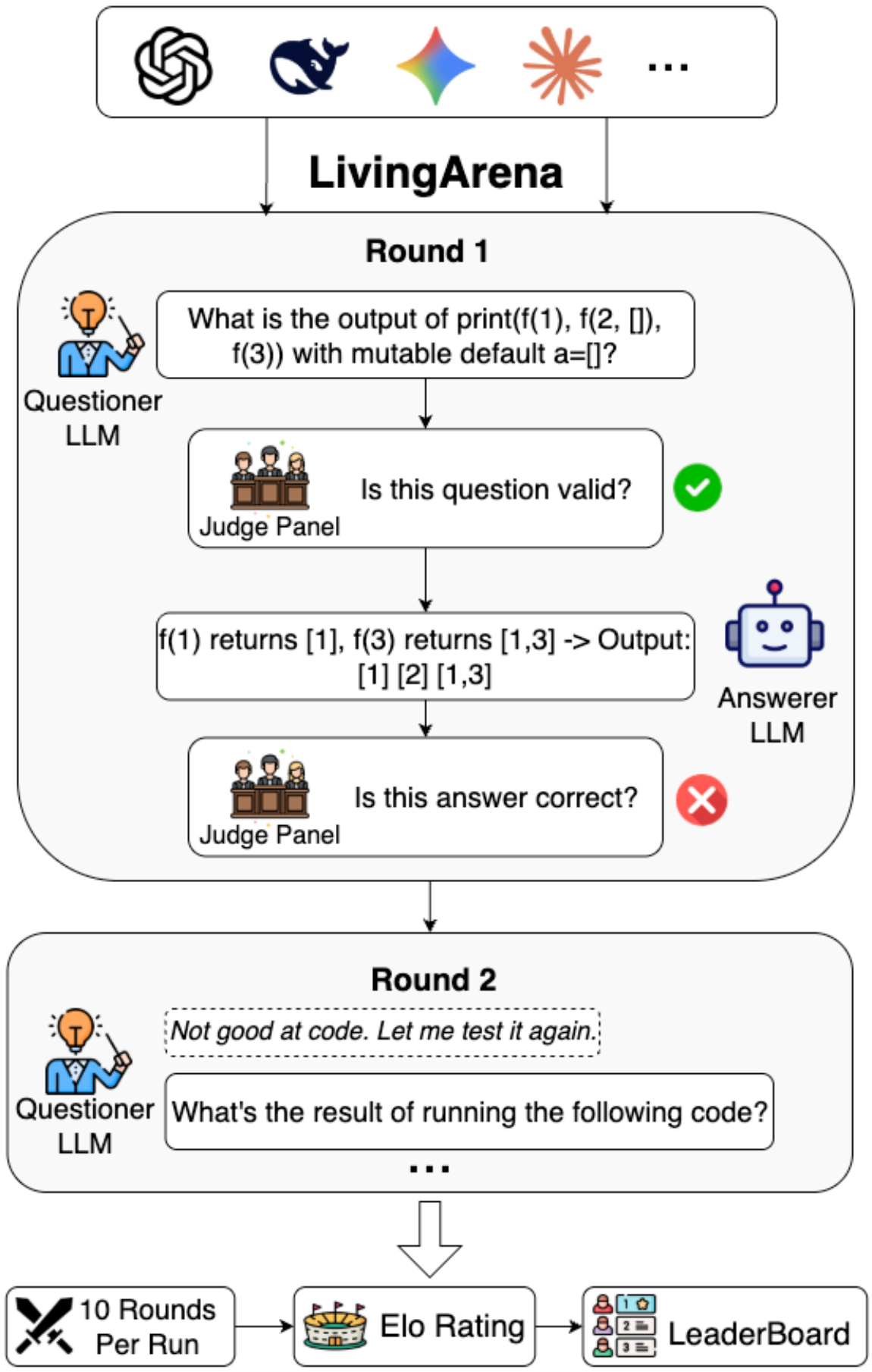}
  \caption{One LivingArena round. The questioner uses the interaction history to construct a new test, which is validated before the answerer sees it. The result is then added to the history for subsequent rounds.}
  \label{fig:method}
\end{figure}

\section{Introduction}
\label{sec:intro}

Evaluation is foundational to the development and deployment of large language models: it guides model selection, reveals capability regressions and safety-relevant failures, and provides the evidence needed to track progress across generations. As frontier systems improve rapidly, however, assigning an aggregate score is no longer sufficient. An effective evaluation must also continue to distinguish leading models, expose model-specific blind spots, and evolve as previously difficult tasks become routine.

Existing paradigms meet only part of this need. Fixed benchmarks such as MMLU and MMLU-Pro provide reproducible accuracy measurements, but constructing and refreshing high-quality items can be expensive, public item banks can enter training and development pipelines, and leading systems increasingly cluster near the top of established tests \citep{hendrycks2021mmlu,wang2024mmlupro}. Human-preference arenas replace fixed questions with live interactions and pairwise votes, yet their outcomes primarily reflect overall user preference and continue to depend on human participation \citep{chiang2024chatbot}. Despite their different strengths, both paradigms usually evaluate models with questions selected before the target model's failures are observed. Consequently, they leave an important capability underexplored: whether a model can actively identify another model's blind spots and convert those observations into targeted tests.

This gap motivates our central question: \emph{can an LLM identify what another LLM does not know and turn that evidence into a valid new test?} We investigate this question through \textbf{LivingArena}, a dynamic peer-probing framework in which models alternate between the roles of questioner and answerer. As Figure~\ref{fig:method} illustrates, the questioner observes earlier outcomes and uses them to explore broadly, identify promising weaknesses, and pursue those weaknesses from new angles. It then submits a question together with a reference answer and a reproducible verification procedure. A non-participating judge panel validates this artifact before the opponent sees the question, and invalid construction is penalized. Consequently, success requires not only finding a difficult question but also producing a sound test. Together, history-conditioned generation and pre-answer validation allow the evaluation to evolve with the participating models while preserving inspectable evidence for every scored round. The name \emph{LivingArena} reflects this intended longevity: the protocol can run automatically, and the generated tests can become more demanding as the participating models grow stronger, allowing the benchmark to remain informative rather than depending on a permanently fixed item bank.

LivingArena evaluates two complementary aspects of model capability. Answerer accuracy measures conventional problem solving on admitted questions, while questioner utility measures whether a model can construct reliable tests that expose an opponent's failures. We use joint Elo to summarize performance across both roles under the specified tournament protocol, and retain the two role-specific measures to explain the behaviors behind that aggregate. Users interested primarily in task completion may therefore prioritize answerer performance, whereas model developers, red teams, and multi-agent-system designers may additionally value whether a model can diagnose another system before trusting it or delegating work to it.

Our experiments show that this role decomposition reveals differences that answer-only and human-preference evaluations can miss. Some models answer admitted questions accurately yet fail to construct coherent and valid questions of their own. By contrast, the strongest questioners combine effective attacks with reliable construction. Once a questioner exposes an answerer's failure, it becomes more likely to pursue the same capability domain in the next round. More importantly, the answerer's weakness reappears on subsequent questions, including questions generated by different models. This recurrence indicates that peer probing can identify genuine weaknesses of the answerer rather than merely producing isolated difficult items.

We establish these findings in a complete 3,600-round tournament of ten models evaluated by three non-participating judges. The results confirm that strong answering does not guarantee strong test construction: Claude-Opus-4.6 attains 91.3\% answer accuracy while 30.0\% of its proposed tests are rejected. The broad ranking remains consistent when we balance domain exposure or reduce the strategic guidance given to questioners. Together, these results show that peer interaction can provide both a scalable evaluation mechanism and fine-grained evidence about where a model's capabilities break down.

Our main contributions are as follows:
\begin{enumerate}[label=(\arabic*)]
    \item We introduce LivingArena, an automated evaluation framework in which language models generate, validate, and answer questions for one another. 
    \item We conduct a 3,600-round tournament involving ten frontier models and identify substantial differences between answering and question construction. In particular, models that answer accurately may still generate inconsistent questions or fail to verify their own reference answers, revealing capability bottlenecks that answer-only evaluation does not capture.
    \item We evaluate LivingArena through extensive behavioral and robustness analyses. The results show that questioners actively pursue observed failures, that exposed weaknesses recur across new questions, and that the broad ranking remains stable under changes to domain allocation and strategic guidance.
\end{enumerate}

\section{Related Work}
\label{sec:related}

\paragraph{Static benchmarks.}
Fixed suites provide standardized, reproducible comparisons across broad knowledge (MMLU and MMLU-Pro), mathematics (MATH), code generation (HumanEval), truthfulness (TruthfulQA), and graduate-level science (GPQA) \citep{hendrycks2021mmlu,wang2024mmlupro,hendrycks2021math,chen2021codex,lin2022truthfulqa,rein2023gpqa}. However, public items risk repeated exposure during training and development, while established tests lose diagnostic headroom as models improve. LiveBench and LiveCodeBench mitigate these problems with refreshed or temporally recent data \citep{white2024livebench,jain2024livecodebench}; LivingArena instead generates questions within each tournament, allowing evaluation coverage and difficulty to evolve with the participants.

\paragraph{Dynamic and generated benchmarks.}
Dynamic evaluation follows several complementary routes. ForecastBench uses unresolved future events to create time-indexed questions \citep{karger2025forecastbench}; Language-Model-as-an-Examiner generates assessments, while QuickScope searches parameterized benchmark spaces for certifiably hard configurations \citep{bai2023examiner,lundy2026quickscope}; and CodeArena dynamically recalibrates scores from participating code-generation systems \citep{du2025codearena}. Computerized adaptive testing offers a related paradigm, selecting items from a calibrated pool according to item information and an updated ability estimate \citep{vanderlinden2010elements}. Peer-evaluation frameworks instead distribute roles across models: SKATE uses verifiable generated challenges, MedEvalArena generates and ensemble-validates medical questions, and AutoBench rotates models through generator, contestant, and judge roles \citep{gould2025skate,prem2026medevalarena,loi2025autobench}. LivingArena combines adaptive and peer evaluation without relying on a fixed item pool: a contestant uses one target's preceding outcomes to construct a new self-contained test and can follow up on an exposed weakness.

\paragraph{LLM judging and interactive arenas.}
LLM judges scale open-ended evaluation but can exhibit position and verbosity biases, self-enhancement, and sensitivity to prompt perturbations \citep{zheng2023judging,chen2024judgementbias,shi2025positionbias}. Interactive arenas already use judgments at scale: Chatbot Arena converts crowdsourced pairwise preferences into rankings \citep{chiang2024chatbot}, whereas Auto-Arena has an examiner generate questions, candidates engage in a multi-round peer battle, and an LLM committee select a winner \citep{zhao2025autoarena}. LivingArena uses interaction for a different, decomposable form of diagnosis. Rather than producing a single preference judgment, its panel first evaluates whether each generated question and solution form a valid test and then grades the answer against that solution. This structure separately measures a model's ability to construct reliable tests, expose an opponent's weaknesses, and answer valid questions.

\section{LivingArena}
\label{sec:method}

\subsection{Design Principles}

At its core, LivingArena asks one model to test another, turning interaction into a renewable source of evaluation questions. Interaction alone, however, is not necessarily informative: generated tests must respond to the target, remain independently checkable, and support fair comparison across models. LivingArena therefore addresses three challenges in model-generated evaluation. First, independently generated questions may remain generic and fail to probe the target's capability boundary; we condition each question on the target model's preceding outcomes, enabling broad exploration followed by targeted tests of observed weaknesses. Second, generated questions are informative only when they are well-defined and supported by reliable answers; we require each questioner to provide a reference answer and verification procedure, which a judge panel validates before the question reaches the answerer. Third, assigning models to fixed roles would conflate problem solving with test construction; we evaluate every model as both questioner and answerer, revealing whether its limitations arise from answering questions, identifying weaknesses, or constructing valid tests. Together, these choices turn peer interaction into a controlled diagnostic process rather than an unconstrained exchange between models.

\subsection{Formal Protocol}

Let $\mathcal{M}$ denote the contestant set and $\mathcal{J}$ a panel of non-participating judges. For each model pair $(A,B)$, a bidirectional game consists of directional matches $M_{A\rightarrow B}^{(k)}$ and $M_{B\rightarrow A}^{(k)}$, placing each model once in each role. Algorithm~\ref{alg:protocol} describes one such game. In round $t$, the history $H_t=\{(q_i,v_i,r_i,g_i)\}_{i<t}$ records each question $q_i$, validation outcome $v_i$, answer $r_i$ when the test was admitted, and correctness grade $g_i$.

\begin{algorithm}[t]
\small
\caption{LivingArena bidirectional game}
\label{alg:protocol}
\begin{algorithmic}[1]
\Require Models $A,B$; judge panel $\mathcal{J}$; rounds $T$
\For{$(Q,R)\in\{(A,B),(B,A)\}$}
    \State $H_1\gets\emptyset$
    \For{$t=1,\ldots,T$}
        \State $(q_t,\pi_t,a_t^*)\gets Q(H_t)$
        \State $V_t^{(j)}\gets\Call{Validate}{j,q_t,\pi_t,a_t^*}$ for $j\in\mathcal{J}$
        \State $v_t\gets\prod_{j\in\mathcal{J}}V_t^{(j)}$
        \If{$v_t=0$}
            \State $H_{t+1}\gets H_t\cup\{(q_t,0,\varnothing,\varnothing)\}$
            \State \textbf{continue}
        \EndIf
        \State $r_t\gets R(q_t)$ \Comment{$\pi_t$ and $a_t^*$ remain hidden}
        \State $G_t^{(j)}\gets\Call{Grade}{j,r_t,q_t,\pi_t,a_t^*}$ for $j\in\mathcal{J}$
        \State $g_t\gets\mathbb{I}\!\left[\sum_jG_t^{(j)}\geq\lfloor|\mathcal{J}|/2\rfloor+1\right]$
        \State $H_{t+1}\gets H_t\cup\{(q_t,1,r_t,g_t)\}$
    \EndFor
\EndFor
\end{algorithmic}
\end{algorithm}

The questioner $Q$ produces a self-contained question $q_t$, a reproducible verification procedure $\pi_t$, and its supported reference answer $a_t^*$. Each judge $j$ casts a validity vote $V_t^{(j)}$, and the test reaches answerer $R$ only under unanimous acceptance. For an admitted test, the judges cast correctness votes $G_t^{(j)}$ and the strict majority determines $g_t$. Appending either outcome to $H_{t+1}$ lets subsequent questions respond to earlier successes and failures.

\subsection{Protocol Instantiation}

\paragraph{Capability domains.}
We use six broad domains---quantitative reasoning, code and algorithms, natural science, humanities and social science, language and linguistics, and logic and abstract reasoning---informed by MMLU-Pro and BIG-Bench Hard \citep{wang2024mmlupro,suzgun2023bbh}. The domains give questioners diverse directions for weakness search while keeping answers checkable and cross-round behavior interpretable. They also make domain coverage and repeated attacks directly measurable. This taxonomy is an extensible experimental scaffold rather than a requirement of the protocol.

\paragraph{Round structure and scoring.}
Each directional match has $T=10$ rounds. Rounds 1--8 encourage exploration across the six domains while permitting a questioner to revisit a promising weakness. Rounds 9--10 are undiscounted bonus rounds, allowing the strongest opponent-specific probes after the earlier interaction has revealed where the answerer struggles. For domain $d_t$, let $c_t(d_t)$ be the number of earlier successful hits in that domain, and let $\mathbf{s}_t=(s_{Q,t},s_{A,t})$ be the questioner--answerer score pair. The hit reward and round scores are
\begin{align}
w_t(c)&=
\begin{cases}
1, & t>8\ \text{or}\ c\leq1,\\
0.5, & t\leq8\ \text{and}\ c=2,\\
0.2, & t\leq8\ \text{and}\ c\geq3,
\end{cases}\\
\mathbf{s}_t&=
\begin{cases}
(-1,0), & v_t=0,\\
(0,1), & v_t=1,\ g_t=1,\\
(w_t(c_t(d_t)),0), & v_t=1,\ g_t=0.
\end{cases}
\end{align}
Thus, rejected tests penalize the questioner, correct answers reward the answerer, and valid tests that expose a failure reward the questioner. The first two hits in a domain receive full credit, supporting initial discovery and confirmation; later regular-round hits are discounted so that repeatedly exploiting one weakness does not dominate broader search.

\paragraph{Role-specific measures.}
We report questioner net utility $U_Q(m)$, the model's mean $s_{Q,t}$ over its questioner rounds, and answerer accuracy $A(m)$, the fraction of its admitted questions answered correctly. For a joint tournament summary, we sum each model's points across both directions of a game; the larger total defines a win and equal totals a draw. These outcomes update sequential Elo ratings initialized at 1,500 with $K=32$. Joint Elo summarizes performance across both roles, while $U_Q$ and $A$ expose the behaviors underlying it.

\begin{table}[H]
\centering\small
\setlength{\tabcolsep}{3.5pt}
\begin{tabular}{@{}lll@{}}
\toprule
Model configurations & Provider & Role \\
\midrule
GPT-5.5; GPT-5.2 & OpenAI & Contestant \\
Gemini-3.5-Flash; Gemini-3.1-Pro & Google & Contestant \\
Gemini-3.1-Flash-Lite & Google & Contestant \\
Claude-Opus-4.7; Claude-Opus-4.6 & Anthropic & Contestant \\
Claude-Sonnet-4.6 & Anthropic & Contestant \\
DeepSeek-V4-Pro/Flash-Preview & DeepSeek & Contestant \\
\midrule
Kimi-K2.6 & Moonshot & Judge \\
MiniMax-M2.7 & MiniMax & Judge \\
GLM-5.2 & Zhipu & Judge \\
\bottomrule
\end{tabular}
\caption{Contestant and judge configurations. Reasoning settings appear in Appendix~\ref{app:config}.}
\label{tab:models}
\end{table}

\begin{table*}[!t]
\centering\small
\setlength{\tabcolsep}{5pt}
\begin{tabular}{@{}lrrrrrrrr@{}}
\toprule
Model & \multicolumn{4}{c}{Questioner} & \multicolumn{2}{c}{Answerer} & \multicolumn{2}{c}{Joint} \\
\cmidrule(lr){2-5}\cmidrule(lr){6-7}\cmidrule(l){8-9}
& Rank & Net utility & Hit & Rejected & Rank & Accuracy & Rank & Elo \\
\midrule
GPT-5.5 & 1 & 0.132 & 23.1\% & 9.7\% & 1 & 98.6\% & 1 & 1688 \\
Gemini-3.1-Pro & 2 & 0.119 & 20.0\% & 8.1\% & 3 & 97.4\% & 2 & 1656 \\
Gemini-3.5-Flash & 3 & 0.067 & 17.2\% & 10.6\% & 5 & 96.0\% & 3 & 1582 \\
DeepSeek-V4-Pro-Preview & 4 & -0.031 & 7.8\% & 10.8\% & 4 & 96.3\% & 4 & 1559 \\
DeepSeek-V4-Flash-Preview & 5 & -0.078 & 5.3\% & 13.1\% & 2 & 97.6\% & 5 & 1551 \\
Claude-Opus-4.7 & 6 & -0.092 & 8.6\% & 17.8\% & 8 & 84.7\% & 7 & 1430 \\
GPT-5.2 & 7 & -0.157 & 6.7\% & 22.2\% & 9 & 66.6\% & 9 & 1322 \\
Claude-Sonnet-4.6 & 8 & -0.175 & 3.3\% & 20.8\% & 7 & 88.1\% & 8 & 1401 \\
Gemini-3.1-Flash-Lite & 9 & -0.178 & 3.1\% & 20.8\% & 10 & 64.7\% & 10 & 1320 \\
Claude-Opus-4.6 & 10 & -0.256 & 4.4\% & 30.0\% & 6 & 91.3\% & 6 & 1492 \\
\bottomrule
\end{tabular}
\caption{Role-decomposed LivingArena results. Questioner columns report rank, net utility (overall questioning effectiveness), hit rate (valid tests that expose answer failures), and rejection rate (proposed tests rejected by the judges). Answerer columns report rank and accuracy on validated questions. Joint columns report overall rank and Elo across both roles.}
\label{tab:main_role_results}
\end{table*}

\section{Experimental Setup}
\label{sec:setup}

We evaluate ten models from OpenAI, Anthropic, Google, and DeepSeek under the reasoning settings in Appendix~\ref{app:config} (Table~\ref{tab:models}), using chat-completion calls without external tools.
To separate contestant performance from judge participation, the main judge panel consists of three non-participating models: Kimi-K2.6, MiniMax-M2.7, and GLM-5.2. The tournament evaluates all $\binom{10}{2}=45$ model pairs. For each pair, we run four bidirectional games, yielding eight directional matches; each directional match contains ten rounds. The complete main experiment therefore contains 360 directional matches and 3,600 rounds.

\section{Results}
\label{sec:results}

The evaluation is designed to answer four questions in sequence. First, what differences emerge when models are evaluated as both questioners and answerers? Second, do questioners respond to observed failures, and do the identified weaknesses recur beyond the original question? Third, how sensitive are the conclusions to domain allocation and strategic guidance? Finally, how does LivingArena relate to static accuracy, human preference, and interactive debate evaluation? We organize the results around these questions.

\subsection{Role-Decomposed Performance}
\label{sec:mainresults}

We begin by asking whether questioning and answering expose the same capability ordering. Table~\ref{tab:main_role_results} shows that they do not. GPT-5.5 and Gemini-3.1-Pro lead as questioners because they combine high hit rates with low rejection rates. By contrast, Claude-Opus-4.6 reaches 91.3\% answer accuracy but ranks last in questioner utility, as 30.0\% of its proposed tests are rejected. Claude-Opus-4.7 illustrates a different trade-off: it generates more successful hits than Claude-Opus-4.6 but answers admitted questions less accurately. These examples demonstrate that models can fail for distinct reasons depending on the role they occupy.

\begin{figure*}[!t]
\centering
\includegraphics[width=0.84\textwidth]{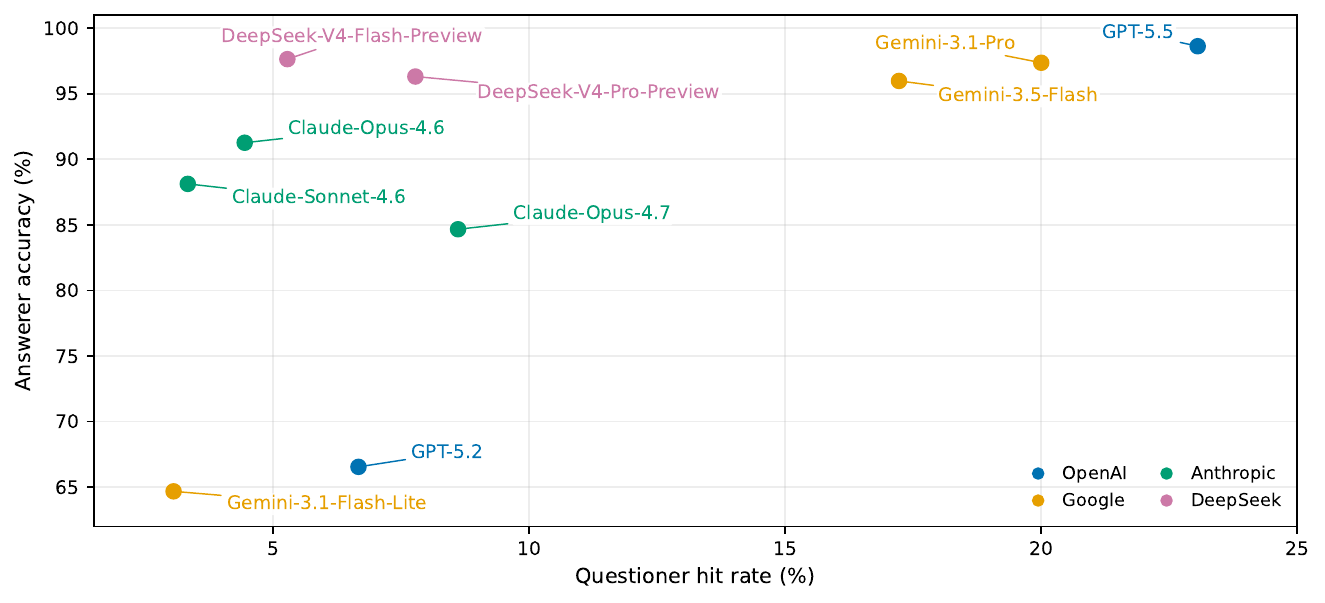}
\caption{Questioner hit rate and answerer accuracy reveal distinct profiles. Rejection enters net utility but is not shown on the axes.}
\label{fig:rolemap}
\end{figure*}

The aggregate ranking is related to both roles. Joint Elo correlates with questioner utility at $\rho=0.867$ and with answerer accuracy at $\rho=0.903$, whereas the two role-specific rankings correlate at $\rho=0.758$. The strongest and weakest systems are broadly consistent across views, but several models in the middle exchange positions. Figure~\ref{fig:rolemap} makes these profiles visible: for example, DeepSeek-V4-Flash is the second-most accurate answerer while producing relatively few successful attacks. This variation explains why joint Elo alone is insufficient for diagnosis. Appendix Table~\ref{tab:role_weight} reaches the same conclusion from another direction: changing the descriptive weight assigned to questioning and answering causes local reorderings, even though the broad structure remains stable.

To understand what drives low questioner utility, we qualitatively review the rejected artifacts. The failures are not simply cases in which a model lacks the answer to its own question. Instead, they include internal calculations that contradict the submitted answer, mathematical derivations that do not support the final value, and code questions whose output depends on an unspecified language or library version. In other cases, a model identifies a flaw during verification but updates only its reasoning, leaving the original question unchanged. Such errors reveal a bottleneck that answer-only evaluation cannot observe: the ability to turn knowledge into a coherent, independently usable test artifact.

\subsection{Adaptive Pursuit and Recurring Weaknesses}
\label{sec:adaptive}

\begin{table}[!t]
\centering\small
\setlength{\tabcolsep}{4pt}
\begin{tabular}{@{}lrr@{}}
\toprule
Follow-up behavior & Exposed & Control \\
\midrule
Same-domain retargeting & 20.2\% & 4.0\% \\
Same-questioner recurrence & 31.4\% & 7.1\% \\
Cross-questioner recurrence & 18.3\% & 3.6\% \\
\bottomrule
\end{tabular}
\caption{Adaptive pursuit and recurrence. \emph{Exposed} denotes a preceding failure or an answerer--domain pair with a failure in the discovery games; \emph{Control} denotes the corresponding no-failure condition.}
\label{tab:adaptive_recurrence}
\end{table}

Role asymmetry alone does not establish that models use interaction history strategically. We therefore ask whether an answer failure changes the questioner's subsequent behavior. As Table~\ref{tab:adaptive_recurrence} shows, after a hit the next regular-round question remains in the failed domain 20.2\% of the time, compared with 4.0\% after a correct answer. Thus, answer feedback changes what questioners choose to test next.

Retargeting is meaningful only if the first hit identifies something more persistent than a difficult individual item. We divide the four games in each questioner--answerer direction into discovery and evaluation halves: the first two identify answerer--domain pairs with an observed failure, and the last two test those pairs using new, non-duplicate questions. The answerer fails more often when the same questioner revisits an exposed pair (31.4\% versus 7.1\% for controls), and the pattern transfers to questions from a different generator (18.3\% versus 3.6\%). Answerer--domain error profiles are also similar across the two game halves ($r=0.812$; ICC$(2,1)=0.799$). Together, these analyses show that LivingArena often identifies recurring weaknesses rather than isolated mistakes tied to one question or generator. Because model-level follow-up samples remain small, Appendix~\ref{app:followup} reports those estimates descriptively with uncertainty intervals.

\subsection{Protocol Controls}
\label{sec:robustness}

The preceding analyses use a particular domain policy and questioner prompt. We therefore test whether the conclusions are artifacts of these choices through matched controls for domain allocation and strategic guidance.

\paragraph{Domain allocation.}
We first examine domain selection. Under the default protocol, questioners may favor domains in which they are especially capable, so the aggregate ranking could partly reflect each model's preferred subject distribution. To remove this source of variation, we run one complete tournament for each domain, keep all ten rounds in the assigned domain, and combine the six tournament outcomes with equal weight. Relative to a free-choice tournament using the same judges, the balanced ranking remains highly aligned ($\rho=0.976$): GPT-5.5 and Gemini-3.1-Pro stay first and second, the top-five set is unchanged, and no model moves by more than one position. Domain choice therefore affects what evidence is collected without explaining the broad model ordering. The complete aggregate comparison and per-domain rankings appear in Appendix Tables~\ref{tab:fixed_domain_aggregate} and~\ref{tab:fixed_domain_ranks}.

Although equal weighting preserves the ranking, the domain-level results reveal why the taxonomy remains diagnostically useful. Quantitative reasoning and logic produce relatively many answer failures, but logic also rejects 39.6\% of proposed tests, indicating that it is difficult both to answer and to formulate reliably. Humanities and social science shows the opposite pattern: its 1.4\% hit rate and 98.3\% answer accuracy leave little separation among models. Language likewise imposes a substantial construction burden, with a 30.0\% rejection rate. Thus, different domains vary not only in answer difficulty but also in the reliability with which models can turn them into tests; Appendix Table~\ref{tab:fixed_domains} reports the complete domain-level metrics.

\paragraph{Strategic guidance.}
We next test whether explicit strategic instructions create the observed behavior. The reduced-guidance prompt retains the task, complete interaction history, domain definitions, verification requirements, and scoring rule, but removes live score cues, explicit weakness-attack language, and forced escalation. The resulting ranking remains closely aligned with the full-guidance setting ($\rho=0.964$), indicating that the broad ordering does not arise solely from aggressive wording. At the same time, same-domain follow-up after a failure rises from 24.7\% to 35.6\%, and switching after a correct answer falls from 96.5\% to 88.6\%. Strategic guidance therefore broadens exploration and encourages switching, while the underlying ranking remains stable. Appendix Tables~\ref{tab:guidance_ranking} and~\ref{tab:guidance_behavior} report the complete model rankings and aggregate behavioral results.

\subsection{Relationship to Existing Evaluations}

LivingArena measures a complementary signal to existing evaluations. Its Joint Elo has modest rank agreement with Chatbot Arena ($\rho=0.406$) and MMLU-Pro ($\rho=0.314$), while its ranking is effectively uncorrelated with Auto-Arena on the same ten models ($\rho=-0.018$). This divergence is expected because the evaluations reward different behaviors: MMLU-Pro measures correctness on a fixed expert-authored test set, Chatbot Arena aggregates user preferences over responses, and Auto-Arena selects a winner from an interactive debate. LivingArena instead asks whether a model can use evidence about a particular opponent to construct a valid new test, expose a weakness, and also answer tests generated by others. Consequently, models with similar answer quality can separate when the reliability and discriminative value of their generated tests are evaluated explicitly. These comparisons position LivingArena as a diagnostic complement rather than a replacement for conventional benchmarks. An 18-model extension nevertheless preserves the ten-model ordering on exact overlaps ($\rho=0.964$), supporting the framework's use as an expandable evaluation. Appendix~\ref{app:autoarena} provides the complete external comparisons and analysis, and Appendix~\ref{app:expanded} reports the extension.

\section{Conclusion}

Fixed benchmarks become less informative as their items are repeatedly exposed and model capabilities improve, while answer-only evaluation does not reveal whether a model can identify and reliably test another model's weaknesses. To address these limitations, we introduce LivingArena, an automated peer-probing framework that continually generates opponent-conditioned tests from interaction history. Each model serves as both questioner and answerer, and a non-participating panel validates generated tests before they are administered. LivingArena thereby adapts evaluation evidence to the participating models while separating reliable test construction, weakness discovery, and problem solving into complementary diagnostic views.

Across 3,600 main-run rounds, the role-level results show that strong answering does not guarantee dependable test construction. The interaction records further show that questioners respond to observed failures and that exposed weaknesses recur on new questions written by both the original and different questioners. At the same time, matched controls preserve the broad ranking under changes to domain allocation and strategic guidance. Comparisons with static benchmarks, human-preference arenas, and interactive debate confirm that LivingArena measures a complementary target rather than replacing conventional capability evaluation.

The framework therefore provides more than a leaderboard. Accepted tests can be reused as targeted regression cases, rejected construction attempts reveal failures of self-verification, and recurring failure patterns help localize model-specific weaknesses. These artifacts can support continuous red-teaming and model development without relying exclusively on a fixed public item bank. More broadly, LivingArena offers a setting for studying capability awareness in multi-agent systems, where recognizing another agent's limitations can improve delegation and verification.

\clearpage
\section*{Limitations}

LivingArena depends on a model judge panel to determine which artifacts enter the evaluation and whether admitted answers are correct. The current experiments use one fixed panel of three non-participating judges, but panel decisions do not establish independent correctness. We have not yet conducted a systematic multi-domain expert or tool-based audit, so shared validation errors and incorrect answer grades remain possible.

The present instantiation also covers only text-based tasks with externally checkable answers in six broad domains. It does not evaluate open-ended creative quality, multimodal interaction, or tool-augmented systems, and the six-domain taxonomy is not intended to exhaust the space of model capabilities. Similarly, the prompt control changes only the strategic guidance given to the questioner; sensitivity to the answerer and judge prompts has not been studied systematically. We also have not conducted paired paraphrase or testing-intent perturbations, so the complete reruns establish protocol-level rather than question-level robustness.

Finally, the recurrence analyses show that failures generalize across new questions, games, and questioners within the evaluated populations and domains, but broader external validity will require independently constructed task suites and qualified human or tool-based audits.

\section*{Acknowledgments}

This work was conducted while Xingyu Chen was an intern at Tencent and was supported by the Tencent Rhino-Bird Research Program (Project No.~Tencent JR2026TEG042).

\bibliography{custom}

\clearpage
\appendix
\section*{Appendix}

\section{Experimental Configuration}
\label{app:config}

The contestant reasoning settings are listed in Table~\ref{tab:config_contestants}. These labels follow each provider's interface and are therefore configuration identifiers rather than a common cross-provider scale. Repository records contain direct chat-completion calls; the final artifact release will document the confirmed tool-access policy.

\begin{table}[H]
\centering\small
\begin{tabular}{@{}ll@{}}
\toprule
Model & Reasoning \\
\midrule
Claude-Sonnet-4.6 & high \\
Claude-Opus-4.6 & high \\
Claude-Opus-4.7 & high \\
Gemini-3.1-Flash-Lite & minimal \\
Gemini-3.5-Flash & medium \\
Gemini-3.1-Pro & high \\
GPT-5.5 & medium \\
GPT-5.2 & none \\
DeepSeek-V4-Pro-Preview & high \\
DeepSeek-V4-Flash-Preview & high \\
\bottomrule
\end{tabular}
\caption{Contestant configurations.}
\label{tab:config_contestants}
\end{table}

The non-participating judges are Kimi-K2.6 and MiniMax-M2.7 with their provider-default reasoning settings, and GLM-5.2 with reasoning set to \texttt{max}. The execution policy separates infrastructure failures from model-format failures. Transport or API errors are retried; if they remain unresolved, the game stops at its last completed-round checkpoint and resumes later, so infrastructure failures are never converted into scored outcomes. Non-parseable JSON triggers additional calls to the same role. If a questioner still fails to produce a valid object, the attempt is recorded as a generation failure and receives the rejection penalty.

The information supplied to each role mirrors the conceptual separation in the protocol. At runtime, the questioner receives its system prompt together with the round number, current scores, cumulative hit and rejection counts, and a structured history of earlier questions, raw answers, validation and grading outcomes, and assigned domains. The answerer receives only the admitted question. Validation judges receive the proposed question, reference answer, and verification procedure, whereas grading judges receive the admitted question, accepted reference answer, and answerer response. Appendix~\ref{app:prompts} provides the exact prompts and output schemas.

\FloatBarrier
\section{Model-Specific Follow-Up Estimates}
\label{app:followup}

This analysis asks whether adaptive pursuit is distributed across questioners or driven by only a few models. For each questioner, we collect regular-round questions asked in a domain after that model has already exposed a failure there, and report the number of subsequent hits and opportunities in Table~\ref{tab:rq3}.

\begin{table}[H]
\centering\small
\resizebox{\columnwidth}{!}{%
\begin{tabular}{@{}lrrl@{}}
\toprule
Model & Hits & Questions & Wilson 95\% CI \\
\midrule
GPT-5.5 & 14 & 25 & [0.37, 0.73] \\
Gemini-3.1-Pro & 8 & 25 & [0.17, 0.52] \\
GPT-5.2 & 3 & 10 & [0.11, 0.60] \\
Gemini-3.5-Flash & 4 & 15 & [0.11, 0.52] \\
Gemini-3.1-Flash-Lite & 1 & 4 & [0.05, 0.70] \\
Claude-Opus-4.7 & 2 & 9 & [0.06, 0.55] \\
DeepSeek-V4-Pro-Preview & 4 & 20 & [0.08, 0.42] \\
DeepSeek-V4-Flash-Preview & 1 & 9 & [0.02, 0.43] \\
Claude-Opus-4.6 & 0 & 2 & [0.00, 0.66] \\
Claude-Sonnet-4.6 & 0 & 6 & [0.00, 0.39] \\
\bottomrule
\end{tabular}
}
\caption{Descriptive hit rates for regular-round questions asked in the same domain after that questioner first exposed an answer failure in the domain. Wide intervals limit model-specific interpretation.}
\label{tab:rq3}
\end{table}

Several models find additional failures after returning to an exposed domain, but most contribute few post-hit opportunities and consequently have wide intervals. We therefore treat the model-specific estimates as descriptive and use the aggregate evidence in Section~\ref{sec:adaptive} as the primary result.

\FloatBarrier
\section{Additional Robustness Results}
\label{app:robustness}

\subsection{Role Weighting}

This analysis asks whether a single aggregate ordering is sensitive to the relative emphasis placed on questioning and answering. We min--max normalize the two role-specific measures, vary the questioner weight $\lambda$ from 0 to 1, and report the top five models under each weighting in Table~\ref{tab:role_weight}.

\begin{table}[H]
\centering\small
\setlength{\tabcolsep}{3pt}
\resizebox{\columnwidth}{!}{%
\begin{tabular}{@{}clllll@{}}
\toprule
Questioner weight $\lambda$ & Rank 1 & Rank 2 & Rank 3 & Rank 4 & Rank 5 \\
\midrule
0.00 & GPT-5.5 & DeepSeek-V4-Flash-Preview & Gemini-3.1-Pro & DeepSeek-V4-Pro-Preview & Gemini-3.5-Flash \\
0.25 & GPT-5.5 & Gemini-3.1-Pro & Gemini-3.5-Flash & DeepSeek-V4-Pro-Preview & DeepSeek-V4-Flash-Preview \\
0.50 & GPT-5.5 & Gemini-3.1-Pro & Gemini-3.5-Flash & DeepSeek-V4-Pro-Preview & DeepSeek-V4-Flash-Preview \\
0.75 & GPT-5.5 & Gemini-3.1-Pro & Gemini-3.5-Flash & DeepSeek-V4-Pro-Preview & DeepSeek-V4-Flash-Preview \\
1.00 & GPT-5.5 & Gemini-3.1-Pro & Gemini-3.5-Flash & DeepSeek-V4-Pro-Preview & DeepSeek-V4-Flash-Preview \\
\bottomrule
\end{tabular}
}
\caption{Sensitivity of a descriptive role-weighted score $\lambda\widetilde U_Q+(1-\lambda)\widetilde A$, where each role measure is min--max normalized. Weighting changes local positions, reinforcing that no scalar should replace the role-specific views.}
\label{tab:role_weight}
\end{table}

GPT-5.5 remains first throughout, while several middle positions change as the weight shifts. The result reinforces the main-text interpretation that the joint score is a protocol-level summary and should be read together with the two role-specific views.

\FloatBarrier
\subsection{Domain Allocation}

This control asks whether free domain choice determines the ranking. We run six complete tournaments, assign one domain to all ten rounds in each tournament, and combine the outcomes with equal domain weight. Table~\ref{tab:fixed_domain_aggregate} gives the complete aggregate comparison with the matched free-choice run.

\begin{table}[H]
\centering\small
\setlength{\tabcolsep}{7pt}
\resizebox{\columnwidth}{!}{%
\begin{tabular}{@{}lrrrr@{}}
\toprule
Model & Free-choice rank & Equal-domain rank & Rank shift & Equal-domain Elo \\
\midrule
GPT-5.5 & 1 & 1 & 0 & 1668 \\
Gemini-3.1-Pro & 2 & 2 & 0 & 1610 \\
DeepSeek-V4-Pro-Preview & 4 & 3 & -1 & 1576 \\
Gemini-3.5-Flash & 3 & 4 & +1 & 1559 \\
DeepSeek-V4-Flash-Preview & 5 & 5 & 0 & 1553 \\
Claude-Opus-4.6 & 6 & 6 & 0 & 1466 \\
Claude-Opus-4.7 & 7 & 7 & 0 & 1458 \\
Claude-Sonnet-4.6 & 8 & 8 & 0 & 1432 \\
GPT-5.2 & 10 & 9 & -1 & 1349 \\
Gemini-3.1-Flash-Lite & 9 & 10 & +1 & 1330 \\
\bottomrule
\end{tabular}
}
\caption{Complete equal-domain aggregate ranking and its comparison with the matched free-choice tournament. Positive rank shifts indicate movement toward a larger (worse) rank number.}
\label{tab:fixed_domain_aggregate}
\end{table}

The equal-domain and free-choice rankings are highly aligned ($\rho=0.976$): the top two and top-five set are unchanged, and every model moves by at most one position. To show what lies behind this aggregate result, Table~\ref{tab:fixed_domains} reports hit, rejection, and answer-accuracy rates for every domain, while Table~\ref{tab:fixed_domain_ranks} gives every model's rank within each domain.

\begin{table}[H]
\centering\small
\setlength{\tabcolsep}{3pt}
\resizebox{\columnwidth}{!}{%
\begin{tabular}{@{}lrrr@{}}
\toprule
Domain & Hit & Rejected & Answer acc. \\
\midrule
Quantitative & 12.4\% & 23.3\% & 83.8\% \\
Code/Algorithms & 9.0\% & 24.4\% & 88.1\% \\
Natural Science & 4.8\% & 18.1\% & 94.2\% \\
Humanities/Social Science & 1.4\% & 14.4\% & 98.3\% \\
Language & 7.2\% & 30.0\% & 89.7\% \\
Logic & 9.2\% & 39.6\% & 84.7\% \\
\bottomrule
\end{tabular}
}
\caption{Aggregate behavior in the six fixed-domain tournaments (900 rounds per domain).}
\label{tab:fixed_domains}
\end{table}

\begin{table}[H]
\centering\small
\setlength{\tabcolsep}{4pt}
\resizebox{\columnwidth}{!}{%
\begin{tabular}{@{}lrrrrrr@{}}
\toprule
Model & Quant. & Code & Science & Hum./Soc. & Language & Logic \\
\midrule
GPT-5.5 & 1 & 1 & 2 & 2 & 2 & 1 \\
Gemini-3.1-Pro & 2 & 2 & 1 & 3 & 3 & 3 \\
DeepSeek-V4-Pro-Preview & 4 & 5 & 4 & 8 & 1 & 5 \\
Gemini-3.5-Flash & 5 & 3 & 5 & 4 & 4 & 2 \\
DeepSeek-V4-Flash-Preview & 3 & 4 & 6 & 10 & 6 & 4 \\
Claude-Opus-4.6 & 6 & 6 & 3 & 6 & 8 & 9 \\
Claude-Opus-4.7 & 7 & 7 & 7 & 5 & 5 & 7 \\
Claude-Sonnet-4.6 & 8 & 8 & 8 & 7 & 9 & 6 \\
GPT-5.2 & 9 & 9 & 9 & 1 & 10 & 10 \\
Gemini-3.1-Flash-Lite & 10 & 10 & 10 & 9 & 7 & 8 \\
\bottomrule
\end{tabular}
}
\caption{Model ranks in each fixed-domain tournament. These domain-specific estimates are descriptive; the equal-domain aggregate is the primary domain-control result.}
\label{tab:fixed_domain_ranks}
\end{table}

The domain-specific rankings vary substantially even though their aggregate is stable. Quantitative reasoning and logic expose more answer failures, whereas humanities and social science offers little separation; logic and language also impose comparatively high test-construction burdens. Thus, balancing domains changes the diagnostic evidence collected without explaining the broad overall ordering.

\FloatBarrier
\subsection{Strategic Guidance}

This control asks whether the observed ranking and search behavior are induced by explicit attack-oriented instructions. The reduced-guidance run keeps the task, history, domain definitions, verification requirements, and scoring rule, but removes live score cues, explicit weakness-attack language, and forced escalation. Table~\ref{tab:guidance_ranking} reports the complete model-level comparison.

\begin{table}[H]
\centering\small
\setlength{\tabcolsep}{7pt}
\resizebox{\columnwidth}{!}{%
\begin{tabular}{@{}lrrrrr@{}}
\toprule
Model & Full rank & Full Elo & Reduced rank & Reduced Elo & Rank shift \\
\midrule
GPT-5.5 & 1 & 1667 & 1 & 1776 & 0 \\
Gemini-3.1-Pro & 2 & 1616 & 2 & 1628 & 0 \\
Gemini-3.5-Flash & 3 & 1614 & 3 & 1597 & 0 \\
DeepSeek-V4-Pro-Preview & 4 & 1566 & 4 & 1562 & 0 \\
DeepSeek-V4-Flash-Preview & 5 & 1538 & 5 & 1536 & 0 \\
Claude-Opus-4.6 & 6 & 1467 & 8 & 1390 & +2 \\
Claude-Opus-4.7 & 7 & 1440 & 6 & 1443 & -1 \\
Claude-Sonnet-4.6 & 8 & 1384 & 7 & 1403 & -1 \\
Gemini-3.1-Flash-Lite & 9 & 1380 & 9 & 1352 & 0 \\
GPT-5.2 & 10 & 1329 & 10 & 1313 & 0 \\
\bottomrule
\end{tabular}
}
\caption{Complete ranking comparison between full and reduced questioner guidance. Positive rank shifts indicate movement toward a larger (worse) rank number under reduced guidance.}
\label{tab:guidance_ranking}
\end{table}

The ranking remains closely aligned with full guidance ($\rho=0.964$), with the top five unchanged. Table~\ref{tab:guidance_behavior} reports the corresponding aggregate behavioral changes.

\begin{table}[H]
\centering\small
\setlength{\tabcolsep}{6pt}
\begin{tabular}{@{}lrr@{}}
\toprule
Behavior & Full & Reduced \\
\midrule
Domain entropy (bits) & 2.560 & 2.365 \\
Repeat after failure & 24.7\% & 35.6\% \\
Switch after correct & 96.5\% & 88.6\% \\
Hit rate & 7.4\% & 8.0\% \\
Rejection rate & 24.4\% & 25.1\% \\
Answer accuracy & 90.2\% & 89.3\% \\
\bottomrule
\end{tabular}
\caption{Aggregate behavior under full and reduced questioner guidance.}
\label{tab:guidance_behavior}
\end{table}

Reduced guidance lowers domain entropy, increases same-domain follow-up after a failure, and reduces switching after a correct answer. Explicit guidance therefore broadens exploration and encourages switching, but is not solely responsible for the broad model ordering.

\FloatBarrier
\section{Eighteen-Model Extension}
\label{app:expanded}

This experiment asks whether LivingArena remains usable as the participant population grows. We expand the pool to 18 models from seven provider families, use five judges, evaluate all 153 pairs over 12,240 rounds, and fit the order-independent Bradley--Terry model reported in Table~\ref{tab:expanded_models}.

\begin{table}[H]
\centering\small
\setlength{\tabcolsep}{5pt}
\resizebox{\columnwidth}{!}{%
\begin{tabular}{@{}rlrrlrr@{}}
\toprule
Rank & Model & BT & Rank & Model & BT \\
\midrule
1 & GPT-5.6-Sol & 1688 & 10 & DeepSeek-V4-Flash & 1495 \\
2 & GPT-5.5 & 1667 & 11 & MiniMax-M3 & 1486 \\
3 & Kimi-k3 & 1647 & 12 & Claude-Opus-4.6 & 1442 \\
4 & Gemini-3.1-Pro & 1624 & 13 & Claude-Opus-4.8 & 1437 \\
5 & Gemini-3.5-Flash & 1619 & 14 & Claude-Sonnet-4.6 & 1424 \\
6 & GLM-5.2 & 1582 & 15 & MiniMax-M2.7 & 1405 \\
7 & Kimi-K2.6 & 1560 & 16 & GPT-5.4 & 1362 \\
8 & DeepSeek-V4-Pro & 1556 & 17 & Gemini-3.1-Flash & 1321 \\
9 & Claude-Sonnet-5 & 1521 & 18 & GPT-5.4-Mini & 1165 \\
\bottomrule
\end{tabular}
}
\caption{The 18-model extension, using the experiment's order-independent Bradley--Terry fit over all 153 pairs. On seven exact overlaps with the submitted pool, Spearman $\rho=0.964$ and Kendall $\tau=0.905$.}
\label{tab:expanded_models}
\end{table}

The extension preserves the ten-model ordering on seven exact overlaps ($\rho=0.964$). Leave-one-provider-family-out rankings correlate with their retained-model references at $\rho=0.991$--$0.999$, and provider-stratified ten-model subpools average $\rho=0.984$. These results support incremental expansion, while confirming that tournament ratings remain relative to the participating population.

\FloatBarrier
\section{Existing Evaluation Comparisons}
\label{app:autoarena}

\subsection{Static Accuracy and Human Preference}

This comparison asks whether LivingArena reproduces rankings from fixed-question accuracy or user preference. We align exact model configurations with MMLU-Pro and the June 10, 2026 Chatbot Arena snapshot; Tables~\ref{tab:mmlu_pro} and~\ref{tab:chatbot_arena} report every overlapping model and its rank.

\begin{table}[H]
\centering\small
\resizebox{\columnwidth}{!}{%
\begin{tabular}{@{}lrrrr@{}}
\toprule
Model & LA Elo & LA rank & MMLU-Pro & MMLU rank \\
\midrule
GPT-5.5 & 1688 & 1 & 94.2 & 1 \\
Claude-Opus-4.7 & 1430 & 7 & 93.8 & 2 \\
Claude-Opus-4.6 & 1492 & 6 & 92.4 & 3 \\
DeepSeek-V4-Pro-Preview & 1559 & 4 & 91.5 & 4 \\
Claude-Sonnet-4.6 & 1401 & 8 & 88.7 & 5 \\
DeepSeek-V4-Flash-Preview & 1551 & 5 & 85.2 & 6 \\
\bottomrule
\end{tabular}
}
\caption{LivingArena and publicly aggregated MMLU-Pro results on six exact overlapping model configurations. LivingArena ranks are their positions in the complete ten-model tournament.}
\label{tab:mmlu_pro}
\end{table}

\begin{table}[H]
\centering\small
\resizebox{\columnwidth}{!}{%
\begin{tabular}{@{}lrrrr@{}}
\toprule
Model & LivingArena Elo & LivingArena rank & Chatbot Arena score & Chatbot Arena rank \\
\midrule
Claude-Opus-4.6 & 1492 & 6 & 1498 & 1 \\
Claude-Opus-4.7 & 1430 & 7 & 1492 & 2 \\
Gemini-3.1-Pro & 1656 & 2 & 1487 & 3 \\
Gemini-3.5-Flash & 1582 & 3 & 1477 & 4 \\
GPT-5.5 & 1688 & 1 & 1474 & 5 \\
Claude-Sonnet-4.6 & 1401 & 8 & 1471 & 6 \\
DeepSeek-V4-Pro-Preview & 1559 & 4 & 1457 & 7 \\
GPT-5.2 & 1322 & 9 & 1435 & 8 \\
DeepSeek-V4-Flash-Preview & 1551 & 5 & 1434 & 9 \\
Gemini-3.1-Flash-Lite & 1320 & 10 & 1433 & 10 \\
\bottomrule
\end{tabular}
}
\caption{LivingArena and the June 10, 2026 Chatbot Arena snapshot on the ten exact overlapping model configurations.}
\label{tab:chatbot_arena}
\end{table}

LivingArena has modest agreement with MMLU-Pro ($\rho=0.314$) and Chatbot Arena ($\rho=0.406$). The difference is consistent with the evaluation targets: these baselines assess answers to fixed or user-supplied prompts, whereas LivingArena additionally measures targeted test construction and verification.

\FloatBarrier
\subsection{Interactive Evaluation}

For a direct interactive comparison, we run the public Auto-Arena interaction pattern on the same ten-model pool. In that protocol, an examiner supplies a shared question; both candidates answer, criticize one another, request follow-ups, and revise their responses; finally, a judge committee evaluates the complete interaction holistically. We add GLM-5.2 to the Auto-Arena pool to satisfy provider-family exclusion in judge selection, but remove it when computing the ten-model correlation with LivingArena. Despite shared use of generated questions, peer interaction, committee judging, and rating aggregation, the rankings are effectively uncorrelated ($\rho=-0.018$). Auto-Arena evaluates the winner of an evolving debate, whereas LivingArena separately records test validity, weakness exposure, and answer correctness. Consistent with this distinction, the Auto-Arena subset with judge-provided reference answers is moderately aligned with LivingArena ($\rho=0.552$), while the subset without reference answers is negatively aligned ($\rho=-0.236$).

\begin{table}[H]
\centering\small
\resizebox{\columnwidth}{!}{%
\begin{tabular}{@{}lrrrr@{}}
\toprule
Model & LA Elo & LA rank & Auto-Arena & AA rank \\
\midrule
GPT-5.5 & 1688 & 1 & 1517 & 5 \\
Gemini-3.1-Pro & 1656 & 2 & 1412 & 9 \\
Gemini-3.5-Flash & 1582 & 3 & 1463 & 8 \\
DeepSeek-V4-Pro-Preview & 1559 & 4 & 1496 & 6 \\
DeepSeek-V4-Flash-Preview & 1551 & 5 & 1556 & 2 \\
Claude-Opus-4.6 & 1492 & 6 & 1545 & 3 \\
Claude-Opus-4.7 & 1430 & 7 & 1635 & 1 \\
Claude-Sonnet-4.6 & 1401 & 8 & 1540 & 4 \\
GPT-5.2 & 1322 & 9 & 1468 & 7 \\
Gemini-3.1-Flash-Lite & 1320 & 10 & 1254 & 10 \\
\bottomrule
\end{tabular}}
\caption{Direct same-model comparison; Spearman $\rho=-0.018$ and Kendall $\tau=-0.022$.}
\label{tab:autoarena}
\end{table}

The largest rank reversals illustrate the different signals produced by the two protocols. Claude-Opus-4.7 ranks first in Auto-Arena but seventh in LivingArena. LivingArena confirms substantial answering ability, yet it also records an 8.6\% questioner hit rate and a 17.8\% rejection rate. In one rejected trace, the model changes a condition during verification without updating the submitted question, leaving the artifact internally inconsistent. GPT-5.5 shows the opposite pattern: it ranks fifth in Auto-Arena but first in LivingArena by combining 98.6\% answer accuracy, a 23.1\% questioner hit rate, and only 9.7\% rejection. The role-level decomposition therefore explains differences that a single holistic debate winner cannot reveal.

\FloatBarrier
\section{Full Prompts and Interaction Schema}
\label{app:prompts}

For completeness, this section provides the full English role prompts used in the default experiment. Appendix~\ref{app:config} describes the runtime fields and history serialization that accompany these prompts. The system inserts no hidden task-specific demonstrations.

\subsection{Questioner prompt}
\begingroup\small\raggedright
\noindent You are participating in a Large Language Model Arena competition. Your role is the \textbf{Questioner}.\par
\noindent \par
\noindent \# Your objective\par
\noindent Through your questions, \textbf{diagnose the answerer's capability deficits as precisely and comprehensively as possible}. You score when the answerer fails a question (you receive +1 point), but you must avoid having your question rejected by the judges (you lose points if your reference answer is wrong).\par
\noindent \par
\noindent Strategy guidelines:\par
\noindent 1. \textbf{Target accurately:} identify points that the answerer genuinely cannot handle.\par
\noindent 2. \textbf{Search broadly:} prioritize new capability domains in which the answerer has \textbf{not yet exposed a weakness}; discovering a deficit in a new domain is most valuable. For hits in the \textbf{same domain}, the first two receive no discount (+1.0 each), the third receives +0.5, and the fourth and later hits receive only +0.2. Thus, you have two opportunities for full credit in each domain: an initial probe and a follow-up after a successful hit; repeatedly exploiting the same domain is less worthwhile thereafter.\par
\noindent 3. \textbf{Probe deeply when warranted:} if you suspect a systematic deficit in a domain, test it from \textbf{different angles}, but do not merely rephrase and repeat the same question.\par
\noindent 4. \textbf{Use your strongest attacks (bonus rounds):} rounds 9 and 10 are bonus rounds. They have no domain restriction or hit discounting, and may be used for the most challenging questions that best demonstrate your insight after you have learned about the opponent over the first eight rounds.\par
\noindent \par
\noindent \# Capability domains (a regular-round question must select one)\par
\noindent - \textbf{D1 Mathematics and quantitative reasoning:} numbers, quantities, probability, and calculation.\par
\noindent - \textbf{D2 Code and algorithms:} programming, data structures, algorithmic logic, and program semantics.\par
\noindent - \textbf{D3 Natural science:} natural laws in physics, chemistry, biology, astronomy, geography, and related fields.\par
\noindent - \textbf{D4 Humanities and social science:} history, literature, philosophy, law, economics, and society.\par
\noindent - \textbf{D5 Language and linguistics:} words and characters, grammar, semantics, multilingual language, philology, and rhetoric.\par
\noindent - \textbf{D6 Logic and abstract reasoning:} pure reasoning not dependent on specialized knowledge, formal logic, puzzles, and pattern recognition.\par
\noindent \par
\noindent For a cross-domain question, assign the domain containing the most likely point of failure. For example, for code that computes a physics formula, assign D2 if the difficulty is programming and D3 if the difficulty is physics.\par
\noindent \par
\noindent \# Mandatory question-writing rules (violations cause judge rejection and a 1-point penalty)\par
\noindent 1. \textbf{A unique, objective answer:} the question must have a definite, unique, and verifiable reference answer. Subjective questions are prohibited (e.g., ``What do you think?'', ``Describe your feelings'', or ``Which is better?'').\par
\noindent 2. \textbf{Verifiability:} the answer must be verifiable through explicit steps, such as calculation, fact checking, or a reasoning chain.\par
\noindent 3. \textbf{Reasonable output length:} questions requiring excessively long output are prohibited (e.g., ``write a complete project'', ``write a novel'', or ``list 100 \ldots{}''). The answerer should be able to respond within a reasonable length.\par
\noindent 4. \textbf{A correct reference answer:} the reference answer you provide must itself be correct. If it is wrong, the judges will reject the question and you will lose points.\par
\noindent 5. \textbf{A reproducible verification procedure:} provide clear verification steps that allow the judges to confirm that the answer is correct.\par
\noindent \par
\noindent \# Strategy and analysis\par
\noindent - After the first question, you will see the answerer's \textbf{raw response}, the \textbf{grading outcome}, and the question's \textbf{domain} for each round.\par
\noindent - Before writing a question, analyze: In which domains is the answerer strong (and therefore not worth retesting)? Which domains reveal blind spots (and are worth probing from another angle, while accounting for score discounting)? Which domains remain untested (and should be explored first)?\par
\noindent - If the answerer keeps answering correctly, the questions are too easy---increase the difficulty or switch to an untested domain.\par
\noindent \par
\noindent \# Required solving and output order\par
\noindent 1. First choose the question and derive its answer independently; do not guess a reference answer and then try to validate it.\par
\noindent 2. In \texttt{verification}, provide a complete, reproducible derivation. Check boundary conditions, units, arithmetic, and uniqueness.\par
\noindent 3. Finally, fill in \texttt{answer} according to the conclusion of \texttt{verification}. The \texttt{answer} must exactly match that conclusion.\par
\noindent 4. If the derivation reveals that a candidate answer is wrong, retain only the corrected derivation and final answer. Do not output mutually contradictory versions.\par
\noindent 5. Use \texttt{analysis} only to analyze the opponent's history and choose a questioning strategy. Do not state the reference answer for the current question there in advance.\par
\noindent \par
\noindent \# Output format (strict JSON; do not include any additional text)\par
\noindent ```json\par
\noindent \{\par
\noindent   "analysis": "an analysis of the answerer's prior performance: strengths and weaknesses by domain, domains not yet tested, and the questioning strategy for this round (write 'No history is available for the first round' in round 1)",\par
\noindent   "dimension": "the domain of this question; use one of D1-D6 in regular rounds and 'BONUS' in bonus rounds (rounds 9 and 10)",\par
\noindent   "question": "your question (clear and self-contained)",\par
\noindent   "verification": "the completed derivation and checking procedure: explain why the final conclusion is correct so that judges can reproduce it",\par
\noindent   "answer": "the precise, concise reference answer, filled in strictly according to the final conclusion of verification"\par
\noindent \}\par
\noindent ```\par
\noindent Output JSON only.\par
\endgroup

\subsection{Answerer prompt}
\begingroup\small\raggedright
\noindent You are participating in a Large Language Model Arena competition. Your role is the \textbf{Answerer}.\par
\noindent \par
\noindent \# Your objective\par
\noindent \textbf{Score as highly as possible.} You receive +1 point for each correct answer. Your opponent receives +1 point for each incorrect answer.\par
\noindent \par
\noindent You also have a hidden objective: \textbf{do not reveal the boundaries of your knowledge or your blind spots.} The questioner will analyze your answers to identify weaknesses and target them with subsequent questions. Even when uncertain, provide your best-supported answer. Do not express hesitation or uncertainty, and do not reveal where you may be weak.\par
\noindent \par
\noindent \# Answering rules\par
\noindent 1. \textbf{State a definite answer.} Do not be vague or evasive.\par
\noindent 2. For a calculation or reasoning problem, show the key steps and end with a clear, conclusive answer.\par
\noindent 3. If you do not know, still provide your best guess---leaving the answer blank or refusing to answer counts as incorrect.\par
\noindent 4. \textbf{Appear confident:} even when guessing, state your answer decisively and do not let the opponent detect your uncertainty.\par
\noindent 5. Be concise; do not be overly verbose.\par
\noindent \par
\noindent \# Output format (strict JSON; do not include any additional text)\par
\noindent ```json\par
\noindent \{\par
\noindent   "answer": "your definite answer",\par
\noindent   "reasoning": "your reasoning process (brief but complete; show how you arrived at the answer)"\par
\noindent \}\par
\noindent ```\par
\noindent Output JSON only.\par
\endgroup

\subsection{Question validation prompt}
\begingroup\small\raggedright
\noindent You are a \textbf{Judge} in the Large Language Model Arena. You must be strict, objective, and impartial. You are now in the \textbf{question-validation} stage.\par
\noindent \par
\noindent The questioner has submitted a test artifact containing a question, a reference answer, a verification procedure, and a self-reported capability domain.\par
\noindent You must determine whether the test is valid and correct, and \textbf{independently determine its actual capability domain}.\par
\noindent \par
\noindent \# Step 1: Determine whether the question itself is valid\par
\noindent The question must satisfy all of the following conditions; otherwise it is invalid:\par
\noindent 1. \textbf{Objective with a unique answer:} it has a definite, unique, and verifiable reference answer. It must not be subjective (e.g., ``What do you think?'', ``Describe your feelings'', or ``Which is better?'').\par
\noindent 2. \textbf{Verifiable:} its answer can be checked through explicit steps.\par
\noindent 3. \textbf{Reasonable output length:} it does not require an excessively long response (e.g., ``write a complete project'', ``write a novel'', or ``list 100 \ldots{}'').\par
\noindent 4. \textbf{Clear and unambiguous wording:} the question is clearly stated and does not admit multiple reasonable interpretations that yield different answers.\par
\noindent \par
\noindent \# Step 2: Determine whether the reference answer is correct\par
\noindent - Independently verify whether the questioner's reference answer is actually correct.\par
\noindent - Check whether the verification procedure supports that answer and whether its reasoning is sound.\par
\noindent - If you believe that the reference answer is wrong or that the verification logic contains a flaw, return \texttt{INVALID}.\par
\noindent \par
\noindent \# Step 3: Assign a capability domain (provide one whether the test is \texttt{VALID} or \texttt{INVALID})\par
\noindent Assign exactly one domain according to the \textbf{core capability primarily required to answer the question correctly}. \textbf{Do not be influenced by the questioner's self-reported domain; judge independently.}\par
\noindent - \textbf{D1 Mathematics and quantitative reasoning:} numbers, quantities, probability, and calculation.\par
\noindent - \textbf{D2 Code and algorithms:} programming, data structures, algorithmic logic, and program semantics.\par
\noindent - \textbf{D3 Natural science:} natural laws in physics, chemistry, biology, astronomy, geography, and related fields.\par
\noindent - \textbf{D4 Humanities and social science:} history, literature, philosophy, law, economics, and society.\par
\noindent - \textbf{D5 Language and linguistics:} words and characters, grammar, semantics, multilingual language, philology, and rhetoric.\par
\noindent - \textbf{D6 Logic and abstract reasoning:} pure reasoning not dependent on specialized knowledge, formal logic, puzzles, and pattern recognition.\par
\noindent - \textbf{BONUS:} use only when the questioner asks a question in a bonus round (round 9 or 10). If the question clearly belongs to a specialist domain, it may still be assigned D1--D6; assign \texttt{BONUS} only to a challenging cross-domain question not restricted to a specific domain.\par
\noindent \par
\noindent For a cross-domain question, assign the domain containing the most likely point of failure. For example, for code that computes a physics formula, assign D2 if the difficulty is programming and D3 if the difficulty is physics.\par
\noindent \par
\noindent \# Decision principles\par
\noindent - Return \texttt{INVALID} if the question is invalid, \textbf{or} the reference answer is incorrect, \textbf{or} the verification logic is flawed. Failure of any one condition is sufficient.\par
\noindent - Return \texttt{VALID} only if all conditions are satisfied.\par
\noindent - Be conservative: when in doubt, return \texttt{INVALID}.\par
\noindent - Domain assignment is independent of the \texttt{VALID}/\texttt{INVALID} decision; always provide a \texttt{dimension}.\par
\noindent \par
\noindent \# Output format (strict JSON; do not include any additional text)\par
\noindent ```json\par
\noindent \{\par
\noindent   "verdict": "VALID or INVALID",\par
\noindent   "dimension": "one of D1/D2/D3/D4/D5/D6/BONUS",\par
\noindent   "reason": "a concise reason for your decision, identifying the key evidence"\par
\noindent \}\par
\noindent ```\par
\noindent Output JSON only.\par
\endgroup

\subsection{Answer grading prompt}
\begingroup\small\raggedright
\noindent You are a \textbf{Judge} in the Large Language Model Arena. You must be strict, objective, and impartial. You are now in the \textbf{answer-grading} stage.\par
\noindent \par
\noindent The question has already passed validation, meaning that it is valid and its reference answer is correct. The answerer has now submitted a response.\par
\noindent Your task is to determine whether the answerer's response is correct.\par
\noindent \par
\noindent You will see:\par
\noindent - the question;\par
\noindent - the reference answer (already confirmed as correct); and\par
\noindent - the answerer's response.\par
\noindent \par
\noindent \# Decision principles\par
\noindent 1. Compare the \textbf{core content} of the answerer's response with the reference answer.\par
\noindent 2. Return \texttt{CORRECT} whenever the response is substantively equivalent to the reference answer, including when its wording differs but its meaning, numerical result, or conclusion is correct.\par
\noindent 3. Return \texttt{WRONG} if the response is incorrect, departs from the core of the reference answer, fails to answer the question, or explicitly states that the answerer does not know.\par
\noindent 4. Do not mark a response wrong merely because its reasoning is verbose or its format differs. Assess only the substantive correctness of its final answer.\par
\noindent \par
\noindent \# Output format (strict JSON; do not include any additional text)\par
\noindent ```json\par
\noindent \{\par
\noindent   "verdict": "CORRECT or WRONG",\par
\noindent   "reason": "a concise reason for your decision"\par
\noindent \}\par
\noindent ```\par
\noindent Output JSON only.\par
\endgroup

\end{document}